\pdfoutput=1

\documentclass[11pt]{article}

\usepackage[]{acl}

\usepackage{times}
\usepackage{latexsym}
\usepackage{multirow}
\usepackage{tabularx}
\usepackage{graphicx}
\usepackage{todonotes}
\usepackage{amssymb} 
\usepackage{booktabs}
\usepackage{comment}
\newcommand{\highlight}[1]{\colorbox{blue!10}{#1}}
\usepackage{hyperref}
\usepackage{scrextend}
\usepackage{tablefootnote}
\usepackage{longtable}
\usepackage[T1]{fontenc}

\usepackage[utf8]{inputenc}

\usepackage{microtype}

\newcommand{\REDSandT}{{\em REDSandT}}
\newcommand{\DISTRE}{{\em DISTRE}}
\newcommand{\RESIDE}{{\em RESIDE}}
\newcommand{\CIL}{{\em CIL}}
\newcommand{\pare}{{\em PARE}}
\newcommand{\mpare}{{\em mPARE}}

\title{\pare: A Simple and Strong Baseline for Monolingual and Multilingual Distantly Supervised Relation Extraction}

\author{Vipul Rathore\footnotemark \hskip 1em  
Kartikeya Badola\footnotemark[\value{footnote}]  \hskip 1em
Parag Singla \hskip 1em Mausam \\
        Indian Institute of Technology \\
  New Delhi, India \\  
   \texttt{rathorevipul28@gmail.com},  \texttt{kartikeya.badola@gmail.com} \\ 
   \texttt{parags@cse.iitd.ac.in},  \texttt{mausam@cse.iitd.ac.in}}

\begin{document}
\maketitle

\begin{abstract}
Neural models for distantly supervised relation extraction (DS-RE) encode each sentence in an entity-pair bag separately. These are then aggregated for bag-level relation prediction. Since, at encoding time, these approaches do not allow information to flow from other sentences in the bag, we believe that they do not utilize the available bag data to the fullest. In response, we explore a simple baseline approach (\pare) in which all sentences of a bag are concatenated into a \emph{passage} of sentences, and encoded jointly using BERT. The contextual embeddings of tokens are aggregated using attention with the candidate relation as query -- this summary of whole passage predicts the candidate relation. We find that our simple baseline solution outperforms existing state-of-the-art DS-RE models in both monolingual and multilingual DS-RE datasets.

\end{abstract}
{\let\thefootnote\relax\footnotetext{{* Equal Contribution}}}
\section{Introduction}
\label{section:intro}

Given some text (typically, a sentence) $t$ mentioning an entity pair $(e_1,e_2)$, the goal of relation extraction (RE) is to predict the relationships between $e_1$ and $e_2$ that can be inferred from $t$. Let $B(e_1,e_2)$ denote the set of all sentences (bag) in the corpus mentioning  $e_1$ and $e_2$ and let $R(e_1,e_2)$ denote all relations from $e_1$ to $e_2$ in a KB. Distant supervision (DS) trains RE models given $B(e_1,e_2)$ and $R(e_1,e_2)$, without sentence level annotation \cite{mintz-etal-2009-distant}. Most DS-RE models use the ``at-least one'' assumption: $\forall r \in R(e_1,e_2)$, $\exists t^r \in B(e_1,e_2)$ such that $t^r$ expresses $(e_1,r,e_2)$.



Recent neural approaches to DS-RE encode each sentence $t\in B(e_1,e_2)$ and then aggregate sentence embeddings using an aggregation operator -- the  common operator being intra-bag attention \cite{lin-etal-2016-neural}. Various models differ in their approach to encoding (e.g., PCNNs, GCNs, BERT) and their loss functions (e.g., contrastive learning, MLM), but agree on the design choice of encoding each sentence independently of the others \cite{vashishth-etal-2018-reside, alt-etal-2019-fine,christou2021improving,chen-etal-2021-cil}. We posit that this choice leads to a suboptimal usage of the available data --  information from other sentences might help in better encoding a given sentence. 

We explore this hypothesis by developing a simple baseline solution. We first construct a \emph{passage} $P(e_1,e_2)$ by concatenating all sentences in $B(e_1,e_2)$. We then encode the whole passage through BERT \cite{devlin-etal-2019-bert} (or mBERT for multilingual setting). This produces a contextualized embedding of every token in the bag. To make these embeddings aware of the candidate relation, we take a (trained) relation query vector, $\mathbf{r}$, to generate a relation-aware summary of the whole passage using attention. This is then used to predict whether $(e_1,r,e_2)$ is a valid prediction. 

\begin{figure*}[!h]
\centering
\includegraphics[width=\textwidth]{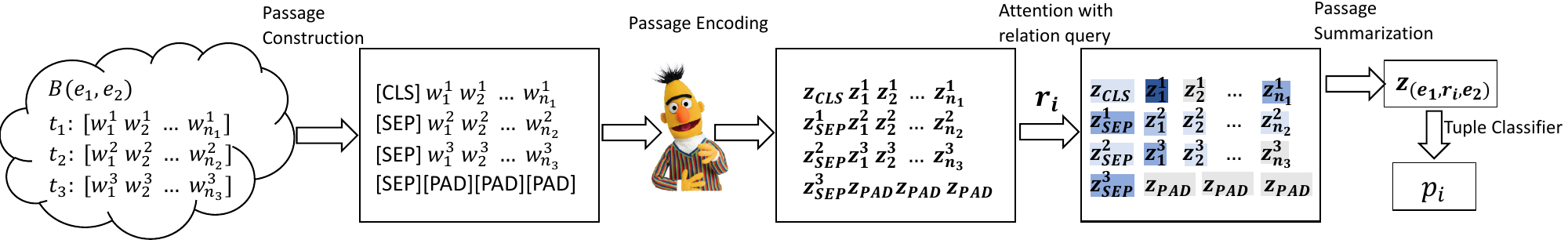}
\caption{Model architecture for \pare. Entity markers not shown for brevity.}
\label{fig:model_Diagram}
\end{figure*}

Despite its simplicity, our baseline has some conceptual advantages. First, each token is able to exchange information with other tokens from other sentences in the bag -- so the embeddings are likely more informed. Second, in principle, the model may be able to relax a part of the at-least-one assumption. For example, if no sentence individually expresses a relation, but if multiple facts in different sentences collectively predict the relation, our model may be able to learn to extract that. 

We name our baseline model Passage-Attended Relation Extraction, \pare\ (\mpare\ for multilingual DS-RE). We experiment on four DS-RE datasets -- three in English, NYT-10d \cite{riedel2010modeling}, NYT-10m, and Wiki-20m \cite{gao2021manual}, and one multilingual, DiS-ReX \cite{bhartiya2021dis}. We find that in all four datasets, our proposed baseline significantly outperforms existing state of the art, yielding up to 5 point AUC gain. Further attention analysis and ablations provide additional insight into model performance. 
We release our code for reproducibility.\footnote{\href{https://github.com/dair-iitd/DSRE}{https://github.com/dair-iitd/DSRE}} We believe that our work represents a simple but strong baseline that can form the basis for further DS-RE research.

\section{Related Work}
\label{section:related_works}
{\bf Monolingual DS-RE:} Early works in DS-RE build probabilistic graphical models for the task (e.g., \cite{hoffmann-etal-2011-knowledge, ritter2013tacl}. Most later works follow the multi-instance multi-label learning framework \cite{surdeanu-etal-2012-multi} in  which there are multiple labels associated with a bag, and the model is trained with at-least-one assumption. Most neural models for the task encode each sentence separately, e.g., using Piecewise CNN \cite{zeng2015distant}, Graph Convolution Net (e.g., \RESIDE\ \cite{vashishth-etal-2018-reside}), GPT (\DISTRE\ \cite{alt-etal-2019-fine}) and BERT (\REDSandT\ \cite{christou2021improving}, \CIL\ \cite{chen-etal-2021-cil}). They all aggregate embeddings using intra-bag attention \cite{lin-etal-2016-neural}. Beyond Binary Cross Entropy, additional loss terms include masked language model pre-training (\DISTRE, \CIL), RL loss \cite{wang-2018-robust}, and auxiliary contrastive learning (\CIL). We show that \pare\ is competitive with \DISTRE, \RESIDE, \CIL, and other natural baselines, without using additional pre-training, side information or auxiliary losses during training, unlike some comparison models.

To evaluate DS-RE, at test time, the model makes a prediction for an unseen bag. Unfortunately, most popular DS-RE dataset (NYT-10d) has a noisy test set, 
as it is automatically annotated \cite{riedel2010modeling}. Recently \citet{gao2021manual} has released NYT-10m and Wiki-20m, which have manually annotated test sets. We use all three datasets in our work.


\vspace{0.5ex}
\noindent{\bf Multilingual DS-RE:} A bilingual DS-RE model named MNRE (tested on English and Mandarin)  
introduced cross-lingual attention in language-specific CNN encoders \cite{lin-etal-2017-neural}.
Recently, \citet{bhartiya2021dis} has released a dataset, DiS-ReX, for four languages -- English, Spanish, German and French. We compare  \mpare\ against the state of the art on DiS-ReX, which combines MNRE architecture with mBERT encoder. See Appendix \ref{sec:appendix_baselines} for details on all
DS-RE models.

\vspace{0.5ex}
\noindent {\bf Passage Construction from Bag of Sentences: } At a high level, our proposed model builds a passage by combining the sentences in a bag that mentions a given entity pair. This idea of passage construction is related with the work of \citet{yan2020bag}, but with important differences, both in task definitions and neural models. First, they focus on predicting the tail entity of a given query $(e_1, r, ?)$, whereas our goal is relation prediction given an entity pair. There are several model differences such as in curating a passage, in use of trainable query vectors for relations, in passage construction strategy, etc.
Importantly, their architecture expects a natural language question for each candidate relation -- not only this requires an additional per-relation annotation (that might not be feasible for datasets having too many relations in the ontology), but also, it makes their method slower, since separate forward passes are needed per relation.

\section{Passage Attended Relation Extraction}
\label{section:model}
\pare\ explores the value of cross-sentence attention during encoding time. It uses a sequence of three key steps: passage construction, encoding and summarization, followed by prediction. Figure \ref{fig:model_Diagram} illustrates these for a three-sentence bag. 

\begin{table*}[h]
\centering

\begin{minipage}{0.26\textwidth}
\centering
\small{\begin{tabular}{@{}lrrrrl@{}}
\toprule
\textbf{Model} & \textbf{AUC} & \textbf{P$@$M} \\ \midrule
 PCNN-Att & 34.1 & 69.4 \\
 RESIDE & 41.5 &  77.2 \\
 DISTRE & 42.2 &  66.8 \\
 REDSandT & 42.4 &  75.3 \\
 CIL & \underline{50.8} &  \highlight{86.0} \\
\textbf{\pare} &  \highlight{53.4} & \underline{84.8}  \\ \bottomrule
\end{tabular}}
\end{minipage}
\begin{minipage}{0.40\textwidth}
\centering
\small{\begin{tabular}{*{7}{c}}
\toprule
\textbf{Model} & \multicolumn{2}{c}{\bf NYT-10m} & \multicolumn{2}{c}{\bf Wiki-20m} \\
  & \textbf{AUC} &  \textbf{M-F1} & \textbf{AUC} & \textbf{M-F1}\\
 \hline
 B+Att & 51.2 &  25.8 & 70.9 &  64.3 \\
 B+Avg & 56.7 &  35.7 & \underline{89.9} &  82.0\\
 B+One & 58.1 & 33.9 & 88.9 &  81.1\\
 CIL & \underline{59.4} & \underline{36.3} & 89.7 & \underline{82.6}\\
 \textbf{\pare} & \highlight{62.2} & \highlight{38.4} & \highlight{91.4} &  \highlight{83.9}\\ \bottomrule
\end{tabular}}
\end{minipage}
\begin{minipage}{0.32\textwidth}
\centering
\small{\begin{tabular}{@{}lrrl@{}}
\toprule
  \textbf{Model} & \textbf{AUC} & \textbf{$\mu$F1} & \textbf{M-F1} \\
 \hline
  PCNN+Att & 67.8 & 63.4 & 43.7 \\
  mB+Att & 80.6 & 74.1 & 69.9 \\
  mB+One & 80.9 & 74.0 & 68.9 \\
  mB+Avg & \underline{82.4} & 75.3 & 71.0 \\
  mB+MNRE & 82.1 & \underline{76.1} & \underline{72.7} \\
  \textbf{\mpare} & \highlight{87.0} & \highlight{79.3} & \highlight{76.0} \\ \bottomrule
\end{tabular}}
\end{minipage}
\caption{Results on (a) NYT-10d, (b) NYT-10m \& Wiki-20m, and (c) DiS-ReX. B=BERT and mB=mBERT. \pare\ and \mpare\ outperforms all models by statistically significant margins (McNemar's test): all $p$ values $<10^{-5}$.}
\label{table:all}
\end{table*}

\vspace{0.5ex}
\noindent
\textbf{Passage Construction} constructs a \emph{passage} $P(e_1,e_2)$ from sentences $t\in B(e_1,e_2)$. The construction process uses a sequential sampling of sentences in the bag without replacement. It terminates if 
(a) adding any new sentence would exceed the maximum number of tokens allowed by the encoder (512 tokens for BERT), or (b) all sentences from the bag have been sampled. 

\vspace{0.5ex}
\noindent
\textbf{Passage Encoding} takes the constructed passage and sends it to an encoder (BERT or mBERT) to generate contextualized embeddings $\mathbf{z}_j$ of every token $w_j$ in the passage. For this, it first creates an encoder input. The input starts with the [CLS] token, followed by each passage sentence separated by [SEP], and pads all remaining tokens with [PAD]. Moreover, following best-practices in RE \cite{han-etal-2019-opennre}, each mention of $e_1$ and $e_2$ in the passage are surrounded by special entity marker tokens <e1>,</e1>, and <e2>,</e2>, respectively. 

\vspace{0.5ex}
\noindent
\textbf{Passage Summarization} maintains a (randomly-initialized) query vector $\mathbf{r}_i$ for every relation $r_i$. It then computes $\alpha_j^i$, the normalized attention of $r_i$ on each token $w_j$, using dot-product attention. Finally, it computes a relation-attended summary of the whole passage $\mathbf{z}_{(e_1,r_i,e_2)}=\sum_{j=1}^{j=L} \alpha_j^i \mathbf{z}_j$, where $L$ is the input length. We note that this summation also aggregates embeddings of [CLS], [SEP], [PAD], as well as entity marker tokens.

\vspace{0.5ex}
\noindent
\textbf{Tuple Classifier} passes  $\mathbf{z}_{(e_1,r_i,e_2)}$ through an MLP followed by Sigmoid activation to return the probability $p_i$ of the triple $(e_1, r_i, e_2)$. This MLP is shared across all relation classes. At inference, a positive prediction is made if $p_i>\mathrm{threshold}$ (0.5).

\vspace{0.5ex}
\noindent
\textbf{Loss Function} is simply Binary Cross Entropy between gold and predicted label set for each bag. No additional loss terms are used.

\section{Experiments and Analysis}
\label{section:experiments}
\begin{figure}[h]
\caption{PR Curve for Models on NYT-10d}
\label{fig:nyt10d}
\centering
\includegraphics[width=7.5cm, height=4cm]{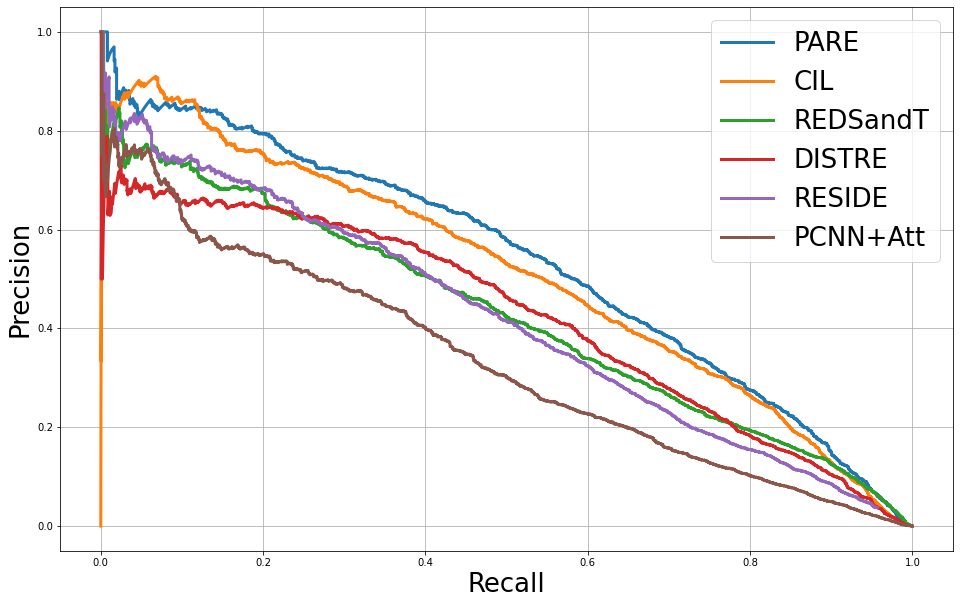}
\end{figure}

\begin{figure}[h]
\caption{PR Curve for Models on DiS-ReX} 
\label{fig:disrex}
\centering
\includegraphics[width=7.5cm,height=4cm]{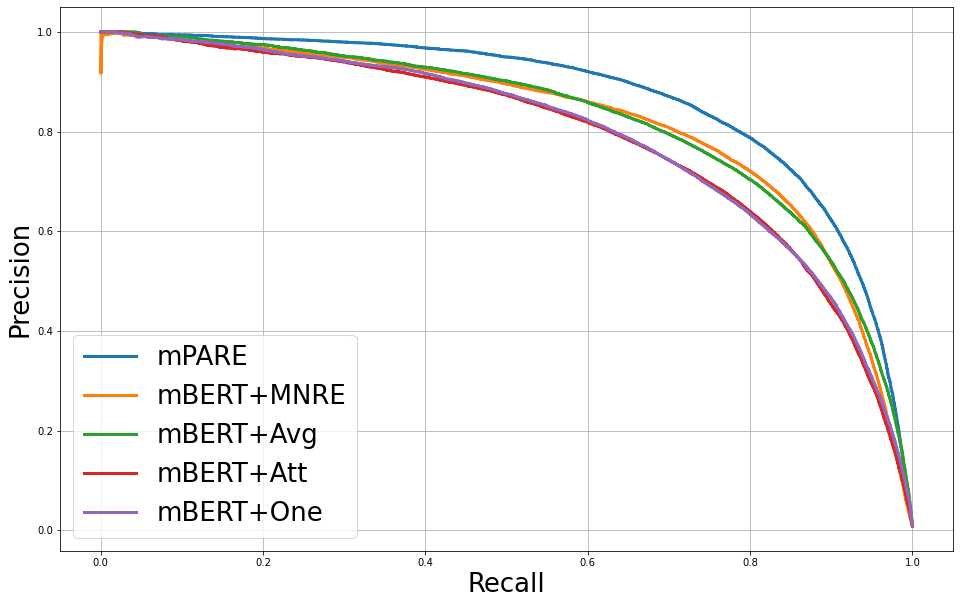}
\end{figure}

We compare \pare\ and \mpare\ against the state of the art models on the respective datasets. We also perform ablations and analyses to understand model behavior and reasons for its performance. 

\vspace{0.5ex}
\noindent
\textbf{Datasets and Evaluation Metrics:}
We evaluate \pare\ on three English datasets: NYT-10d, NYT-10m, Wiki-20m. 
\mpare\ is compared using the DiS-ReX benchmark. Data statistics are in Table \ref{table:stats}, with more details in Appendix \ref{sec:appendix_datastats}. We use the evaluation metrics prevalent in literature for each dataset. These include AUC: area under the precision-recall curve, M-F1: macro-F1, $\mu$-F1: micro-F1, and $P@M$: average of P@100, P@200 and P@300, where P@k denotes precision calculated over a model's $k$ most confidently predicted triples. 


\vspace{0.5ex}
\noindent
\textbf{Comparison Models and Hyperparameters:} Since there is substantial body of work on NYT-10d, we compare against several recent models: \RESIDE, \DISTRE, \REDSandT\ and the latest state of the art, \CIL. For NYT-10m and Wiki-20m, we report comparisons against models in the original paper \cite{gao2021manual}, and also additionally run CIL for a stronger comparison. For DiS-ReX, we compare against mBERT based models. See Appendix \ref{sec:appendix_baselines} for more details on the baseline models. For \pare\ and \mpare, we use base-uncased checkpoints for BERT and mBERT, respectively. Hyperparameters are set based on a simple grid search over devsets. (see Appendix \ref{sec:appendix_settings}).

\begin{table}[t]
\centering
\small{\begin{tabular}{@{}lrrrl@{}}
\toprule
  \textbf{Dataset} & \textbf{\#Rels} & \textbf{\#Total} & \textbf{\#Test} & \textbf{Test set} \\ 
 \hline
 NYT-10d & 58 & 694k & 172k & Distant Sup. \\ 
 NYT-10m & 25 & 474k & 9.74k & Manual \\
 Wiki-20m & 81 & 901k & 140k & Manual \\
 DiS-ReX & 37 & 1.84M & 334k & Distant Sup. \\ \bottomrule
\end{tabular}}
\caption{Dataset statistics.}
\label{table:stats}
\end{table}

\subsection{Comparisons against State of the Art}

The results are presented in Table \ref{table:all}, in which, the best numbers are highlighted and second best numbers are underlined.
On NYT-10d (Table \ref{table:all}(a)), \pare\ has 2.6 pt AUC improvement over \CIL, the current state of the art, while achieving slightly lower P@M. This is also reflected in the P-R curve (Figure \ref{fig:nyt10d}), where in the beginning our P-R curve is slightly on the lower side of CIL, but overtakes it for higher threshold values of recall. Our model beats \REDSandT\ by 11 AUC pts, even though both use BERT, and latter uses extra side-information (e.g., entity-type, sub-tree parse). 

On manually annotated testsets (Table \ref{table:all}(b)), \pare\ achieves up to 2.8 pt AUC and 2.1 pt macro-F1 gains against \CIL. We note that \citet{gao2021manual} only published numbers on simpler baselines (BERT followed by attention, average and max aggregators, the details for which can be found in Appendix \ref{sec:appendix_baselines}), which are substantially outperformed by \pare. \CIL's better performance is likely attributed to its contrastive learning objective -- it will be interesting to study this in the context of \pare. 

For multilingual DS-RE (Table \ref{table:all}(c)), \mpare\ obtains a 4.9 pt AUC gain against mBERT+MNRE. P-R curve in Figure \ref{fig:disrex} shows that it convincingly outperforms others across the entire domain of recall values. We provide language-wise and relation-wise metrics in Appendix \ref{sec:appendix_disrex} -- the gains are consistent on all languages and nearly all relations. 

\subsection{Analysis and Ablations}

\textbf{Generalizing to Unseen KB:} Recently, \citet{ribeiro2020beyond} has proposed a robustness study in which entity names in a bag are replaced by other names (from the same type) to test whether the extractor is indeed reading the text, or is simply overfitting on the regularities of the given KB. We also implement a similar robustness study (details in Appendix \ref{sec:appendix_ood}), where entity replacement results in an entity-pair bag that does not exist in the original KB. We find that on this modified NYT-10m, all models suffer a drop in performance, suggesting that models are not as robust as we intend them to be. We, however, note that \CIL\ suffers a 28.1\% drop in AUC performance, but \pare\ remains more robust with only a 16.8\% drop. We hypothesize that this may be because of \pare's design choice of attending on all words for a given relation, which could reduce its focus on entity names themselves.

\begin{figure}[t]
\caption{AUC on different bins of the NYT-10m test set. The x-axis denotes the range of lengths of untruncated passages in each bin}
\label{fig:bin_main}
\centering
\includegraphics[width=7.5cm, height=5cm]{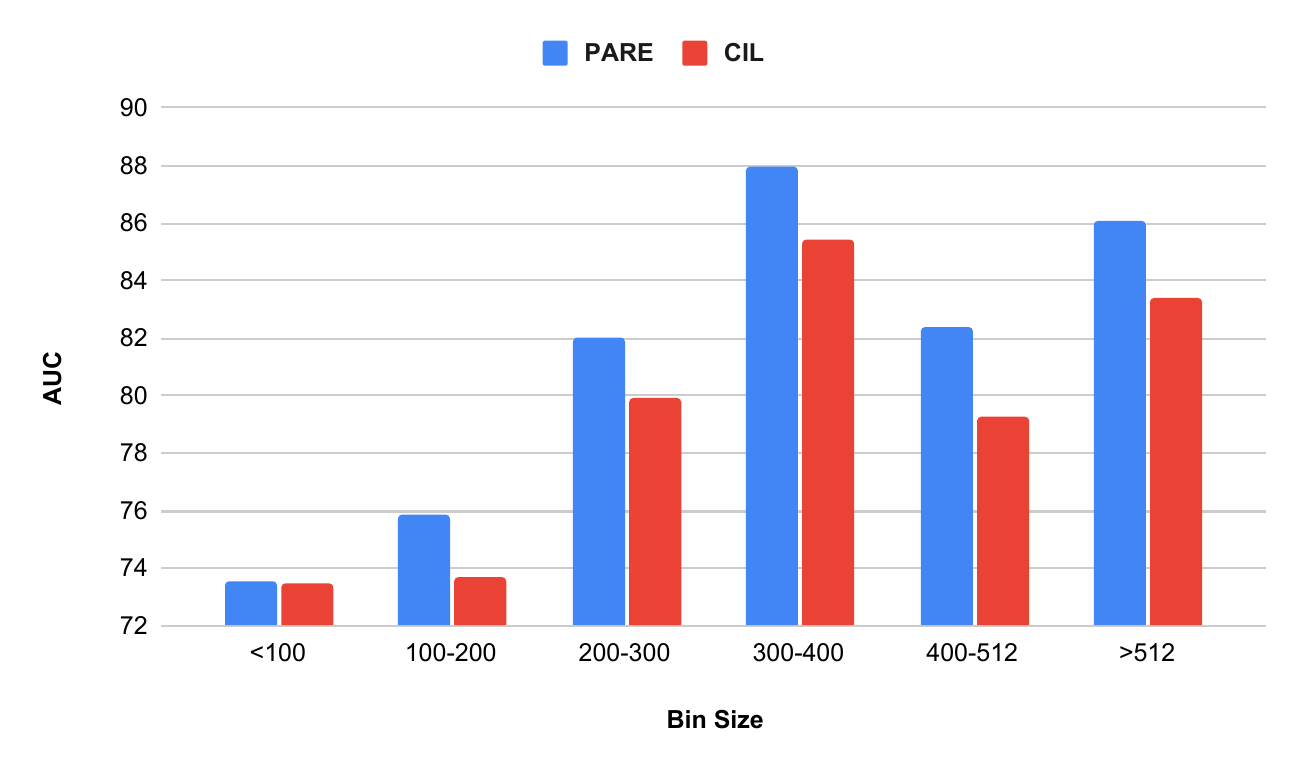}
\end{figure}
\vspace{0.5ex}
\noindent
\textbf{Scaling with Size of Entity-Pair Bags:} Due to truncation when the number of tokens in a bag exceed 512 (limit for BERT), one would assume that \pare\  may not be suited for cases where the number of tokens in a bag is large. To study this, we divide the test set of NYT-10m into 6 different bins based on the number of tokens present in the untruncated passage (details on the experiment in Appendix \ref{sec:appendix_len}). We present results in Figure \ref{fig:bin_main}. We find that \pare\ shows consistent gains of around 2 to 3 pt in AUC against \CIL\ for all groups except the smallest group. This is not surprising, since for smallest group, there is likely only one sentence in a bag, and \pare\ would not gain from inter-sentence attention. For large bags, relevant information is likely already present in truncated passage, due to redundancy.

\vspace{0.5ex}
\noindent
\textbf{Attention Patterns:}
In \pare, each relation class has a trainable query vector, which  attends on every token. The attention scores could give us some insight about the words the model is focusing on. We observe that for a candidate relation that is not a gold label for a particular bag, surprisingly, the highest attention scores are obtained by  [PAD] tokens. In fact, for such bags, on an average, roughly 90\% of the attention weight goes to [PAD] tokens, 
whereas this number is only 0.1\% when the relation is in the gold set (see Appendices \ref{sec:appendix_pad} and \ref{sec:appendix_attwt}). We find this to be an example of model ingenuity -- \pare\ seems to have creatively learned that whenever the most appropriate words for a relation are not present, it could simply attend on [PAD] embeddings, which may lead to similar attended summaries, which may be easily decoded to a low probability of tuple validity. In fact, as a further test, we perform an ablation where we disallow relation query vectors to attend on [PAD] tokens -- this results in an over 3 pt  drop in AUC on NYT-10d, indicating the importance of padding for prediction (see Table \ref{table:ablation_main}).

\begin{table}[t]
\centering
\begin{tabular}{@{}lr@{}}
\toprule
  \textbf{Modification} & \textbf{Change in AUC}\\
 \hline
w/o passage summarization & -4.9 \\
w/o [PAD] attention & -3.1 \\
w/o entity markers & -36.9 \\
\bottomrule
\end{tabular}
\caption{Change in AUC on NYT-10d by removing various architectural components from \pare}
\label{table:ablation_main}
\end{table}
\vspace{0.5ex}
\noindent
\textbf{Ablations:} We perform further ablations of the model by removing entity markers and removing the relation-attention step that computes a summary (instead using [CLS] token for predicting each relation). \pare\ loses significantly in performance in each ablation obtaining 16.5 and 48.5 AUC, respectively (as against 53.4 for full model) on NYT-10d (table \ref{table:ablation_main}). The critical importance of entity markers is not surprising, since without them the model does not know what is the entity-pair it is predicting for. We also notice a very significant gain due to relation attention and passage summarization, suggesting that this is an important step for the model -- it allows focus on specific words relevant for predicting a relation. We perform the same experiments on the remaining datasets and observe similar results (Appendix \ref{sec:appendix_ablation}). 

\vspace{0.5ex}
\noindent
\textbf{Effect of Sentence Order:} 
We build 20 random passages per bag (by varying sentence order and also which sentences get selected if passage needs truncation). On all four datasets (Appendix \ref{sec:appendix_random}), we find that the standard deviation to be negligible. This analysis highlights 1) the sentence-order invariance of \pare's performance and 2) In practical settings, the randomly sampled sentences with token limit of 512 in the passage is good enough to make accurate bag-level predictions.

\section{Conclusion and Future Work}
We introduce \pare, a simple baseline for the task of distantly supervised relation extraction. Our experiments demonstrate that this simple baseline produces very strong results for the task, and outperforms existing top models by varying margins across four datasets in monolingual and multilingual settings. Several experiments for studying model behavior show its consistent performance that generalizes across settings. We posit that our framework would serve as a strong backbone for further research in the field of DS-RE.

There are several directions to develop the \pare\ architecture further. E.g., \pare\ initializes
relation embeddings randomly and also constructs passage via random sampling. Alternatively, one could make use of label descriptions and aliases from Wikidata to initialize label query vectors; one could also use a sampling strategy to filter away noisy sentences  (e.g. a sentence selector \cite{wang-2018-robust} module integrated with \pare). In the multilingual setting, contextualized embeddings of entity mentions in a passage may be aligned using constrained learning techniques \cite{mehta2018towards,NandwaniPMS19} to learn potentially better token embeddings. 
Constraints can be imposed on the label hierarchy as well (E.g. \textit{PresidentOf} $\Rightarrow$ \textit{CitizenOf}, etc.) since label query vectors operate independently of each other on the passage in \pare.  Additionally, translation-based approaches at training or inference \cite{nag2021data, kolluru-acl22} could improve \mpare\ performance. Recent ideas of joint entity and relation alignment in multilingual KBs \cite{singh-akbc21} may be combined along with \mpare's relation extraction capabilities.


\section*{Acknowledgements}
 This work is primarily supported by a grant from Huwaei. Mausam is supported by grants from Google and Jai Gupta Chair Professorship. Parag was supported by the DARPA Explainable Artificial Intelligence (XAI) Program (\#N66001-17-2-4032). Mausam and Parag are supported by IBM SUR awards. Vipul is supported by Prime Minister's Research Fellowship (PMRF). We thank IIT Delhi HPC facility for compute resources. We thank Abhyuday Bhartiya for helping in reproducing results from the DiS-ReX paper, and Keshav Kolluru for helpful comments on an earlier draft of the paper. Any opinions, findings, conclusions or recommendations expressed here are those of the authors and do not necessarily reflect the views or official policies, either expressed or implied, of the funding agencies.

\bibliography{anthology}

\begin{thebibliography}{32}
\expandafter\ifx\csname natexlab\endcsname\relax\def\natexlab#1{#1}\fi

\bibitem[{Alt et~al.(2019)Alt, H{\"u}bner, and Hennig}]{alt-etal-2019-fine}
Christoph Alt, Marc H{\"u}bner, and Leonhard Hennig. 2019.
\newblock \href {https://doi.org/10.18653/v1/P19-1134} {Fine-tuning pre-trained
  transformer language models to distantly supervised relation extraction}.
\newblock In \emph{Proceedings of the 57th Annual Meeting of the Association
  for Computational Linguistics}, pages 1388--1398, Florence, Italy.
  Association for Computational Linguistics.

\bibitem[{Bhartiya et~al.(2022)Bhartiya, Badola, and Mausam}]{bhartiya2021dis}
Abhyuday Bhartiya, Kartikeya Badola, and Mausam. 2022.
\newblock {DiS-ReX}: A multilingual dataset for distantly supervised relation
  extraction.
\newblock In \emph{Proceedings of the 60th Annual Meeting of the Association
  for Computational Linguistics}, Dublin, Ireland. Association for
  Computational Linguistics.

\bibitem[{Chen et~al.(2021)Chen, Shi, Tang, Chen, Wu, and
  Zhuang}]{chen-etal-2021-cil}
Tao Chen, Haizhou Shi, Siliang Tang, Zhigang Chen, Fei Wu, and Yueting Zhuang.
  2021.
\newblock \href {https://doi.org/10.18653/v1/2021.acl-long.483} {{CIL}:
  Contrastive instance learning framework for distantly supervised relation
  extraction}.
\newblock In \emph{Proceedings of the 59th Annual Meeting of the Association
  for Computational Linguistics and the 11th International Joint Conference on
  Natural Language Processing (Volume 1: Long Papers)}, pages 6191--6200,
  Online. Association for Computational Linguistics.

\bibitem[{Christou and Tsoumakas(2021)}]{christou2021improving}
Despina Christou and Grigorios Tsoumakas. 2021.
\newblock Improving distantly-supervised relation extraction through bert-based
  label and instance embeddings.
\newblock \emph{IEEE Access}, 9:62574--62582.

\bibitem[{Devlin et~al.(2019)Devlin, Chang, Lee, and
  Toutanova}]{devlin-etal-2019-bert}
Jacob Devlin, Ming-Wei Chang, Kenton Lee, and Kristina Toutanova. 2019.
\newblock \href {https://doi.org/10.18653/v1/N19-1423} {{BERT}: Pre-training of
  deep bidirectional transformers for language understanding}.
\newblock In \emph{Proceedings of the 2019 Conference of the North {A}merican
  Chapter of the Association for Computational Linguistics: Human Language
  Technologies, Volume 1 (Long and Short Papers)}, pages 4171--4186,
  Minneapolis, Minnesota. Association for Computational Linguistics.

\bibitem[{Gao et~al.(2021)Gao, Han, Bai, Qiu, Xie, Lin, Liu, Li, Sun, and
  Zhou}]{gao2021manual}
Tianyu Gao, Xu~Han, Yuzhuo Bai, Keyue Qiu, Zhiyu Xie, Yankai Lin, Zhiyuan Liu,
  Peng Li, Maosong Sun, and Jie Zhou. 2021.
\newblock \href {https://doi.org/10.18653/v1/2021.findings-acl.112} {Manual
  evaluation matters: Reviewing test protocols of distantly supervised relation
  extraction}.
\newblock In \emph{Findings of the Association for Computational Linguistics:
  ACL-IJCNLP 2021}, pages 1306--1318, Online. Association for Computational
  Linguistics.

\bibitem[{Han et~al.(2020)Han, Gao, Lin, Peng, Yang, Xiao, Liu, Li, Zhou, and
  Sun}]{han-etal-2020-data}
Xu~Han, Tianyu Gao, Yankai Lin, Hao Peng, Yaoliang Yang, Chaojun Xiao, Zhiyuan
  Liu, Peng Li, Jie Zhou, and Maosong Sun. 2020.
\newblock \href {https://aclanthology.org/2020.aacl-main.75} {More data, more
  relations, more context and more openness: A review and outlook for relation
  extraction}.
\newblock In \emph{Proceedings of the 1st Conference of the Asia-Pacific
  Chapter of the Association for Computational Linguistics and the 10th
  International Joint Conference on Natural Language Processing}, pages
  745--758, Suzhou, China. Association for Computational Linguistics.

\bibitem[{Han et~al.(2019)Han, Gao, Yao, Ye, Liu, and
  Sun}]{han-etal-2019-opennre}
Xu~Han, Tianyu Gao, Yuan Yao, Deming Ye, Zhiyuan Liu, and Maosong Sun. 2019.
\newblock \href {https://doi.org/10.18653/v1/D19-3029} {{O}pen{NRE}: An open
  and extensible toolkit for neural relation extraction}.
\newblock In \emph{Proceedings of EMNLP-IJCNLP: System Demonstrations}, pages
  169--174.

\bibitem[{Hoffmann et~al.(2011)Hoffmann, Zhang, Ling, Zettlemoyer, and
  Weld}]{hoffmann-etal-2011-knowledge}
Raphael Hoffmann, Congle Zhang, Xiao Ling, Luke Zettlemoyer, and Daniel~S.
  Weld. 2011.
\newblock \href {https://aclanthology.org/P11-1055} {Knowledge-based weak
  supervision for information extraction of overlapping relations}.
\newblock In \emph{Proceedings of the 49th Annual Meeting of the Association
  for Computational Linguistics: Human Language Technologies}, pages 541--550,
  Portland, Oregon, USA. Association for Computational Linguistics.

\bibitem[{Kingma and Ba(2015)}]{kingma2014adam}
Diederik~P. Kingma and Jimmy Ba. 2015.
\newblock \href {http://arxiv.org/abs/1412.6980} {Adam: {A} method for
  stochastic optimization}.
\newblock In \emph{3rd International Conference on Learning Representations,
  {ICLR} 2015, San Diego, CA, USA, May 7-9, 2015, Conference Track
  Proceedings}.

\bibitem[{Kolluru et~al.(2022)Kolluru, Muqeeth, Mittal, Chakrabarti, and
  Mausam}]{kolluru-acl22}
Keshav Kolluru, Mohammed Muqeeth, Shubham Mittal, Soumen Chakrabarti, and
  Mausam. 2022.
\newblock {A}lignment-{A}ugmented {C}onsistent {T}ranslation for {M}ultilingual
  {O}pen {I}nformation {E}xtraction.
\newblock In \emph{Proceedings of the 60th Annual Meeting of the Association
  for Computational Linguistics}, Dublin, Ireland. Association for
  Computational Linguistics.

\bibitem[{Lin et~al.(2017)Lin, Liu, and Sun}]{lin-etal-2017-neural}
Yankai Lin, Zhiyuan Liu, and Maosong Sun. 2017.
\newblock \href {https://doi.org/10.18653/v1/P17-1004} {Neural relation
  extraction with multi-lingual attention}.
\newblock In \emph{Proceedings of the 55th Annual Meeting of the Association
  for Computational Linguistics (Volume 1: Long Papers)}, pages 34--43,
  Vancouver, Canada. Association for Computational Linguistics.

\bibitem[{Lin et~al.(2016)Lin, Shen, Liu, Luan, and Sun}]{lin-etal-2016-neural}
Yankai Lin, Shiqi Shen, Zhiyuan Liu, Huanbo Luan, and Maosong Sun. 2016.
\newblock \href {https://doi.org/10.18653/v1/P16-1200} {Neural relation
  extraction with selective attention over instances}.
\newblock In \emph{Proceedings of the 54th Annual Meeting of the Association
  for Computational Linguistics (Volume 1: Long Papers)}, pages 2124--2133,
  Berlin, Germany. Association for Computational Linguistics.

\bibitem[{Loshchilov and Hutter(2019)}]{loshchilov2017decoupled}
Ilya Loshchilov and Frank Hutter. 2019.
\newblock \href {https://openreview.net/forum?id=Bkg6RiCqY7} {Decoupled weight
  decay regularization}.
\newblock In \emph{International Conference on Learning Representations}.

\bibitem[{McNemar(1947)}]{mcnemar1947note}
Quinn McNemar. 1947.
\newblock Note on the sampling error of the difference between correlated
  proportions or percentages.
\newblock \emph{Psychometrika}, 12(2):153--157.

\bibitem[{Mehta et~al.(2018)Mehta, Lee, and Carbonell}]{mehta2018towards}
Sanket~Vaibhav Mehta, Jay~Yoon Lee, and Jaime~G Carbonell. 2018.
\newblock Towards semi-supervised learning for deep semantic role labeling.
\newblock In \emph{Proceedings of the 2018 Conference on Empirical Methods in
  Natural Language Processing}, pages 4958--4963.

\bibitem[{Mintz et~al.(2009)Mintz, Bills, Snow, and
  Jurafsky}]{mintz-etal-2009-distant}
Mike Mintz, Steven Bills, Rion Snow, and Daniel Jurafsky. 2009.
\newblock \href {https://aclanthology.org/P09-1113} {Distant supervision for
  relation extraction without labeled data}.
\newblock In \emph{Proceedings of the Joint Conference of the 47th Annual
  Meeting of the {ACL} and the 4th International Joint Conference on Natural
  Language Processing of the {AFNLP}}, pages 1003--1011, Suntec, Singapore.
  Association for Computational Linguistics.

\bibitem[{Nag et~al.(2021)Nag, Samanta, Mukherjee, Ganguly, and
  Chakrabarti}]{nag2021data}
Arijit Nag, Bidisha Samanta, Animesh Mukherjee, Niloy Ganguly, and Soumen
  Chakrabarti. 2021.
\newblock A data bootstrapping recipe for low-resource multilingual relation
  classification.
\newblock In \emph{Proceedings of the 25th Conference on Computational Natural
  Language Learning}, pages 575--587.

\bibitem[{Nandwani et~al.(2019)Nandwani, Pathak, Mausam, and
  Singla}]{NandwaniPMS19}
Yatin Nandwani, Abhishek Pathak, Mausam, and Parag Singla. 2019.
\newblock \href
  {https://proceedings.neurips.cc/paper/2019/hash/cf708fc1decf0337aded484f8f4519ae-Abstract.html}
  {A primal dual formulation for deep learning with constraints}.
\newblock In \emph{Advances in Neural Information Processing Systems 32: Annual
  Conference on Neural Information Processing Systems 2019, NeurIPS 2019,
  December 8-14, 2019, Vancouver, BC, Canada}, pages 12157--12168.

\bibitem[{Paszke et~al.(2019)Paszke, Gross, Massa, Lerer, Bradbury, Chanan,
  Killeen, Lin, Gimelshein, Antiga, Desmaison, Kopf, Yang, DeVito, Raison,
  Tejani, Chilamkurthy, Steiner, Fang, Bai, and Chintala}]{NEURIPS2019_9015}
Adam Paszke, Sam Gross, Francisco Massa, Adam Lerer, James Bradbury, Gregory
  Chanan, Trevor Killeen, Zeming Lin, Natalia Gimelshein, Luca Antiga, Alban
  Desmaison, Andreas Kopf, Edward Yang, Zachary DeVito, Martin Raison, Alykhan
  Tejani, Sasank Chilamkurthy, Benoit Steiner, Lu~Fang, Junjie Bai, and Soumith
  Chintala. 2019.
\newblock \href
  {http://papers.neurips.cc/paper/9015-pytorch-an-imperative-style-high-performance-deep-learning-library.pdf}
  {Pytorch: An imperative style, high-performance deep learning library}.
\newblock In H.~Wallach, H.~Larochelle, A.~Beygelzimer, F.~d\textquotesingle
  Alch\'{e}-Buc, E.~Fox, and R.~Garnett, editors, \emph{Advances in Neural
  Information Processing Systems 32}, pages 8024--8035. Curran Associates, Inc.

\bibitem[{Qin et~al.(2018)Qin, Xu, and Wang}]{wang-2018-robust}
Pengda Qin, Weiran Xu, and William~Yang Wang. 2018.
\newblock Robust distant supervision relation extraction via deep reinforcement
  learning.
\newblock In \emph{Proceedings of the 56th Annual Meeting of the Association
  for Computational Linguistics, {ACL} 2018, Melbourne, Australia, July 15-20,
  2018, Volume 1: Long Papers}, pages 2137--2147. Association for Computational
  Linguistics.

\bibitem[{Radford et~al.(2018)Radford, Narasimhan, Salimans, and
  Sutskever}]{radford2018improving}
Alec Radford, Karthik Narasimhan, Tim Salimans, and Ilya Sutskever. 2018.
\newblock Improving language understanding by generative pre-training.

\bibitem[{Ribeiro et~al.(2020)Ribeiro, Wu, Guestrin, and
  Singh}]{ribeiro2020beyond}
Marco~Tulio Ribeiro, Tongshuang Wu, Carlos Guestrin, and Sameer Singh. 2020.
\newblock Beyond accuracy: Behavioral testing of nlp models with checklist.
\newblock In \emph{Proceedings of the 58th Annual Meeting of the Association
  for Computational Linguistics}, pages 4902--4912.

\bibitem[{Riedel et~al.(2010)Riedel, Yao, and McCallum}]{riedel2010modeling}
Sebastian Riedel, Limin Yao, and Andrew McCallum. 2010.
\newblock Modeling relations and their mentions without labeled text.
\newblock In \emph{Joint European Conference on Machine Learning and Knowledge
  Discovery in Databases}, pages 148--163. Springer.

\bibitem[{Ritter et~al.(2013)Ritter, Zettlemoyer, Mausam, and
  Etzioni}]{ritter2013tacl}
Alan Ritter, Luke Zettlemoyer, Mausam, and Oren Etzioni. 2013.
\newblock Modeling missing data in distant supervision for information
  extraction.
\newblock \emph{Trans. Assoc. Comput. Linguistics}, 1:367--378.

\bibitem[{Singh et~al.(2021)Singh, Chakrabarti, Jain, Choudhury, and
  Mausam}]{singh-akbc21}
Harkanwar Singh, Soumen Chakrabarti, Prachi Jain, Sharod~Roy Choudhury, and
  Mausam. 2021.
\newblock Multilingual knowledge graph completion with joint relation and
  entity alignment.
\newblock In \emph{3rd Conference on Automated Knowledge Base Construction,
  {AKBC} 2021, Virtual, October 4-8, 2021}.

\bibitem[{Soares et~al.(2019)Soares, Fitzgerald, Ling, and
  Kwiatkowski}]{soares2019matching}
Livio~Baldini Soares, Nicholas Fitzgerald, Jeffrey Ling, and Tom Kwiatkowski.
  2019.
\newblock Matching the blanks: Distributional similarity for relation learning.
\newblock In \emph{Proceedings of the 57th Annual Meeting of the Association
  for Computational Linguistics}, pages 2895--2905.

\bibitem[{Surdeanu et~al.(2012)Surdeanu, Tibshirani, Nallapati, and
  Manning}]{surdeanu-etal-2012-multi}
Mihai Surdeanu, Julie Tibshirani, Ramesh Nallapati, and Christopher~D. Manning.
  2012.
\newblock \href {https://aclanthology.org/D12-1042} {Multi-instance multi-label
  learning for relation extraction}.
\newblock In \emph{Proceedings of the 2012 Joint Conference on Empirical
  Methods in Natural Language Processing and Computational Natural Language
  Learning}, pages 455--465, Jeju Island, Korea. Association for Computational
  Linguistics.

\bibitem[{Vashishth et~al.(2018)Vashishth, Joshi, Prayaga, Bhattacharyya, and
  Talukdar}]{vashishth-etal-2018-reside}
Shikhar Vashishth, Rishabh Joshi, Sai~Suman Prayaga, Chiranjib Bhattacharyya,
  and Partha Talukdar. 2018.
\newblock \href {https://doi.org/10.18653/v1/D18-1157} {{RESIDE}: Improving
  distantly-supervised neural relation extraction using side information}.
\newblock In \emph{Proceedings of the 2018 Conference on Empirical Methods in
  Natural Language Processing}, pages 1257--1266, Brussels, Belgium.
  Association for Computational Linguistics.

\bibitem[{Wolf et~al.(2020)Wolf, Debut, Sanh, Chaumond, Delangue, Moi, Cistac,
  Rault, Louf, Funtowicz et~al.}]{wolf2019huggingface}
Thomas Wolf, Lysandre Debut, Victor Sanh, Julien Chaumond, Clement Delangue,
  Anthony Moi, Pierric Cistac, Tim Rault, R{\'e}mi Louf, Morgan Funtowicz,
  et~al. 2020.
\newblock Transformers: State-of-the-art natural language processing.
\newblock In \emph{Proceedings of the 2020 conference on empirical methods in
  natural language processing: system demonstrations}, pages 38--45.

\bibitem[{Yan et~al.(2020)Yan, Han, Sun, Liu, and Bian}]{yan2020bag}
Lingyong Yan, Xianpei Han, Le~Sun, Fangchao Liu, and Ning Bian. 2020.
\newblock From bag of sentences to document: Distantly supervised relation
  extraction via machine reading comprehension.
\newblock \emph{arXiv preprint arXiv:2012.04334}.

\bibitem[{Zeng et~al.(2015)Zeng, Liu, Chen, and Zhao}]{zeng2015distant}
Daojian Zeng, Kang Liu, Yubo Chen, and Jun Zhao. 2015.
\newblock Distant supervision for relation extraction via piecewise
  convolutional neural networks.
\newblock In \emph{Proceedings of the 2015 conference on empirical methods in
  natural language processing}, pages 1753--1762.

\end{thebibliography}
\bibliographystyle{acl_natbib}
\newpage

\appendix
\onecolumn
\section{Experimental Settings}
\label{sec:appendix_settings}
We train and test our model on two NVIDIA GeForce GTX 1080 Ti cards. We use a linear LR scheduler having weight decay of 1e-5 with AdamW \citep{loshchilov2017decoupled,kingma2014adam} as the optimizer. Our implementation uses PyTorch \citep{NEURIPS2019_9015}, the Transformers library \citep{wolf2019huggingface} and OpenNRE \footnote{\href{https://github.com/thunlp/OpenNRE}{https://github.com/thunlp/OpenNRE}} \citep{han-etal-2019-opennre}. We use \href{https://huggingface.co/bert-base-uncased}{bert-base-uncased} checkpoint for BERT initialization in the mono-lingual setting. For multi-lingual setting, we use \href{https://huggingface.co/bert-base-multilingual-uncased}{bert-base-multilingual-uncased}. 

For hyperparameter tuning, we perform grid search over \{1e-5, 2e-5\} for learning rate and \{16, 32, 64\} for batch size and select the best performing configuration for each dataset.

\pare\ takes 2 epochs to converge on NYT-10d (152 mins/epoch), 3 epochs for NYT-10m (138 mins/epoch), 2 epochs for Wiki-20m (166 mins/epoch) and 4 epochs for DiS-ReX (220 mins/epoch).

The numbers we report for the baselines come from their respective papers. We obtained the code base of CIL, BERT+Att, BERT+Avg, BERT+One from their respective authors, so that we could run them on additional datasets. We were able to replicate same numbers as reported in their papers. We trained those models on other datasets as well by carefully tuning the bag size hyperparameter. 
\section{Sizes of different models}
We report the number of additional trainable parameters, in each model, on top of the underlying BERT/mBERT encoder (all models except MNRE use the \href{https://huggingface.co/bert-base-uncased}{bert-base-uncased} checkpoint, whereas MNRE uses the \href{https://huggingface.co/bert-base-multilingual-uncased}{bert-base-multilingual-uncased} checkpoint) in table \ref{tab:num_params}. We note that the key reason why \pare\ has significantly lower number of additional parameters (on top of the BERT/mBERT encoder) is because all the other models use \emph{entity pooling} \cite{soares2019matching} for constructing instance representations. The \emph{entity pooling} operation requires an additional fully-connected layer which projects the concatenated encoded representations of head and tail entity in an input instance to a vector of the same size (for BERT/mBERT, this results in additional $(2\times768)^2$ weight and $2\times768$ bias parameters).


\label{sec:appendix_params}
\begin{table}[h!]
\centering
\begin{tabular}{@{}lr@{}}
\toprule
\textbf{Model} & \textbf{\#Parameters (excluding BERT)} \\ \midrule
Att      & 2400793        \\
One      & 2399257        \\
Avg      & 2399257            \\
CIL      & 2453052   \\
MNRE & 2645029 \\
\textbf{\pare} & 46082   \\ \bottomrule
\end{tabular}
\caption{Comparison of trainable parameters between our model and other state-of-the-art models}
\label{tab:num_params}
\end{table}

\section{Dataset Details}
\label{sec:appendix_datastats}
We evaluate our proposed model on four different datasets: NYT-10d \cite{riedel2010modeling}, NYT-10m \cite{gao2021manual}, Wiki-20m \cite{gao2021manual} and DiS-ReX \cite{bhartiya2021dis}. The statistics for each of the datasets is present in table \ref{table:stats}.
\\ \\
\textbf{NYT-10d}\\
NYT-10d is the most popular dataset for monolingual DS-RE, constructed by aligning Freebase entities to the New York Times Corpus. The train and test splits are both distantly supervised.\\ \\
\textbf{NYT-10m}\\
NYT-10m is a recently released dataset to train and evaluate models for monolingual DS-RE. The dataset is built from the same New York Times Corpus and the Freebase KB but with a new relation ontology and a manually annotated test set. It aims to tackle the existing problems with the NYT-10d dataset by 1) establishing a public validation set 2) establishing consistency among the relation classes present in the train and test set 3) providing a high quality, manually labeled test set.\\ \\
\textbf{Wiki-20m}\\
Wiki-20m is also a recently released dataset for training DS-RE models and evaluating them on manually annotated a test set. The test set in this case corresponds to the Wiki80 dataset \cite{han-etal-2019-opennre}. The relation ontology of Wiki80 is used to re-structure the Wiki20 DS-RE dataset \cite{han-etal-2020-data}, from which the training and validation splits are created. It is made sure that their is no overlap between the instances present in the testing and the training and validation sets.\\ \\ 
\textbf{DiS-ReX}\\
DiS-ReX is a recently released benchmarking dataset for training and evaluating DS-RE models on instances spanning multiple languages. The entities present in this dataset are linked across the different languages which means that a bag can contain sentences from more than one languages. We use the publicly available train, validation and test splits and there is no overlap between the bags present in any two different dataset splits. 
\\
We obtain the first three datasets from \href{https://github.com/thunlp/OpenNRE#datasets}{OpenNRE} and DiS-ReX from their \href{https://github.com/dair-iitd/DiS-ReX}{official repository}.

\section{Description of Intra-Bag attention}
\label{sec:intra_bag}
Let $t_1, t_2, ... , t_n $ denote $n$ instances sampled from $B(e_1,e_2)$. In all models using intra-bag attention for instance-aggregation, each $t_i$ is independently encoded to form the instance representation, $E(t_i)$, following which the relation triple representation $B_r$ for the triple $(e_1, e_2, r)$ is given by $B_r = \sum_{i=0}^{i=n} \alpha^r_i E(t_i)$. Here $r$ is any one of the relation classes present in the dataset and $\alpha^r_i$ is the normalized attention score allotted to instance representation $E(t_i)$ by relation query vector  $\overrightarrow{r}$ for relation $r$. The model then predicts whether the relation triple is a valid one by sending each $B_r$ through a feed-forward neural network. In some variants,  $\overrightarrow{r}$ is replaced with a shared query vector for all relation-classes, $\overrightarrow{q}$, resulting in a bag-representation $B$ corresponding to $(e_1,e_2)$ as opposed to triple-representation.
\section{Baselines}
\label{sec:appendix_baselines}
The details for each baseline is provided below:\\ \\
\textbf{PCNN-Att} \\ 
\citet{lin-etal-2016-neural} proposed the intra-bag attention aggregation scheme in 2016, obtaining the then state-of-the-art performance on NYT-10d using a piecewise convolutional neural network (PCNN \cite{zeng2015distant}).
\\ \\
\textbf{RESIDE} \\ 
\citet{vashishth-etal-2018-reside} proposed RESIDE which uses side-information (in the form of entity types and relational aliases) in addition to sentences present in the dataset. The model uses intra-bag attention with a shared query vector to combine the representations of each instance in the bag. The sentence representations are obtained using a Graph Convolutional Network (GCN) encoder. \\ \\
\textbf{DISTRE} \\ 
\citet{alt-etal-2019-fine} propose the use of a pre-trained transformer based language model (OpenAI GPT \citet{radford2018improving}) for the task of DS-RE. The model uses intra-bag attention for the instance aggregation step.  \\ \\
\textbf{REDSandT} \\
\citet{christou2021improving} propose the use of a BERT encoder for DS-RE by using sub-tree parse of the input sentence along with special entity type markers for the entity mentions in the text. The model uses intra-bag attention for the instance aggregation step. \\ \\
\textbf{CIL} \\
\citet{chen-etal-2021-cil} propose the use of Masked Language Modeling (MLM) and Contrastive Learning (CL) losses as auxilliary losses to train a BERT encoder + Intra-bag attention aggregator for the task. \\ \\ 
\textbf{BERT+Att/mBERT+Att} \\ 
The model uses intra-bag attention aggregator on top of a BERT/mBERT encoder. \\ \\
\textbf{BERT+Avg/mBERT+Avg} \\ 
The model uses ``Average'' aggregator which weighs each instance representation uniformly, hence denoting bag-representation as the average of instance-representations. \\ \\
\textbf{BERT+One/mBERT+One} \\
The model independently performs multi-label classification on each instance present in the bag and then aggregates the classification results by performing class-wise max-pooling (over sentence scores). In essence, the ``One'' aggregator ends up picking one instance for each class (the one which denotes the highest confidence for that particular class), hence the name. \\ \\ 
\textbf{mBERT+MNRE} \\ 
The MNRE aggregator was originally introduced by \citet{lin-etal-2017-neural} and used with a shared mBERT encoder by \citet{bhartiya2021dis} \footnote{Obtained from the \href{https://github.com/dair-iitd/DiS-ReX}{original repository} for DiS-ReX}. The model assigns a query vector for each $(relation, language)$ tuple. A bag is divided into sub-bags where each sub-bag contains the instances of the same language. In essence, a bag has $L$ sub-bags and each relation class corresponds to $L$ query vectors, where $L$ denotes the number of languages present in the dataset. These are then used to construct $L^2$ triple representations (using intra-bag attention aggregation on each \emph{(sub-bag,query vector)} pair for a candidate relation) which are then scored independently. The final confidence score for a triple is the average of $L^2$ triple scores. \\ \\ 

\section{Statistical Significance}
\label{sec:appendix_sign}
We compare the predictions of our model on the non-NA triples present in the test set with the predictions of the second-best model using the McNemar's test of statistical significance \cite{mcnemar1947note}.  
In all cases, we obtained the \emph{p-value} to be many orders of magnitude smaller than 0.05, suggesting that the improvement in results is statistically significant in all cases. 

\section{Ablation Study} 
\label{sec:appendix_ablation}

\begin{table}[h!]
\centering
\begin{tabular}{@{}lrrrr@{}}
\toprule
  \textbf{Modification} & \textbf{NYT-10d} & \textbf{NYT-10m} & \textbf{Wiki-20m} & \textbf{DiS-ReX} \\
 \hline
w/o passage summarization & -4.9 & -2.9 & -4.2 & -0.8\\
w/o [PAD] attention & -3.1 & -2.3 & -1.9 & -0.1\\
w/o entity markers & -36.9 & -16.5 & -29.9 & -20.5\\
\bottomrule
\end{tabular}
\caption{Model ablation i.e. change in AUC performance with different components of \textit{PARE}}
\label{table:ablation}
\end{table}

We perform ablation studies on various datasets to understand which components are most beneficial for our proposed model. We provide the results in table \ref{table:ablation}. 


We observe that upon replacing our passage summarization step with multi-label classification using [CLS] token (present at the start of the passage), we observe a significant decrease in AUC, indicating that contextual embedding of [CLS] token might not contain enough information for multi-label prediction of bag.

For NYT-10, it is interesting to note here that the AUC is still higher than that of REDSandT, a model which uses BERT+Att as the backbone (along with other complicated machinery). This means that one can simply obtain an improvement in performance by creating a passage from multiple instances in a bag.

Removing entity markers resulted in the most significant drop in performance. However, this is also expected since without them, our model would have no way to understand which entities to consider while performing relation extraction.

\section{Attention on [PAD] tokens}
\label{sec:appendix_pad}
In the passage summarization step (described in section \ref{section:model}), we allow the relation query vector $\overrightarrow{r}$ to also attend over the encodings of the [PAD] tokens present in the passage. We make this architectural choice in-order to provide some structure to the relation-specific summaries created by our model. If a particular relation class $r$ is not a valid relation for entity pair $(e_1,e_2)$, then ideally, we would want the attended-summary of the passage $P(e_1,e_2)$ created by the relation vector $\overrightarrow{r}$ to represent some sort of a null state (since information specific to that relation class is not present in the passage). Allowing [PAD] tokens to be a part of the attention would provide enough flexibility to the model to represent such a state. We test our hypothesis by considering 1000 non-NA bags correctly labelled by our trained model in the test set of NYT-10d. Let $R(e_1,e_2)$ denote the set of valid relation-classes for entity pair $(e_1,e_2)$ and let $R$ denote all of the relation-classes present in the dataset. We first calculate the percentage of attention given to [PAD] tokens for a given passage $P(e_1,e_2)$ for all relation-classes in $R$. The results are condensed into two scores, sum of scores for $R(e_1,e_2)$ and sum of scores for $R\setminus R(e_1,e_2)$. The results are aggregated for all 1000 bags, and then averaged out by dividing with the total number of positive triples and negative triples respectively. We obtain that on an average, only 0.07\% of attention weight is given to [PAD] tokens by relation vectors corresponding to $R(e_1,e_2)$, compared to 88.35\% attention weight given by relation vectors corresponding to $R \setminus R(e_1,e_2)$. We obtain similar statistics on other datasets as well. This suggests that for invalid triples, passage summaries generated by the model resemble the embeddings of the [PAD] token. Furthermore, since we don't allow [PAD] tokens to be a part of self-attention update inside BERT, the [PAD] embeddings at the output of the BERT encoder are not dependent on the passage, allowing for uniformity across all bags.

Finally, we train a model where we don't allow the relation query vectors to attend on the [PAD] token embeddings and notice a 3.1pt drop in AUC on NYT-10d (table \ref{table:ablation}). We also note that the performance is still significantly higher than models such as REDSandT and DISTRE, suggesting that our instance aggregation scheme still performs better than the baselines, even when not optimized fully.

\section{Examples of Attention Weighting during Passage Summarization}
\label{sec:appendix_attwt}
To understand how the query vector of a relation attends over passage tokens to correctly predict that relation, we randomly selected from correctly predicted non-NA triples and selected the token obtaining the highest attention score (by the query vector for the correct relation). For the selection, we ignore the stop words, special tokens and the entity mentions. The results are presented in table \ref{tab:att_viz}. 

\begin{table}[ht]
    \centering
    \small{\begin{tabular}{p{0.5\linewidth} | p{0.32\linewidth}}
      \textbf{Input Passage (tokenized by BERT)}  & \textbf{correctly predicted label} \\ \hline
      [CLS] six months later , his widow met the multi \#\#mill \#\#ion \#\#aire [unused2] vincent astor [unused3] , a \textbf{descendant} of the fur trader turned manhattan real - estate magnate [unused0] john jacob astor [unused1] , and a man considered so unpleasant by his peers \- l \#\#rb \- and even by his own mother \- rr \#\#b - that he reportedly required a solitary seating for lunch at his club because nobody would share a meal with him . [SEP] & /people/person/children  \\ \hline
      [CLS] the [unused2] robin hood foundation [unused3] , \textbf{founded} by [unused0] paul tudor jones [unused1] ii and perhaps the best - known hedge fund charity , raised \$ 48 million at its annual benefit dinner last year . [SEP]      & /business/person/company        \\ \hline
      [CLS] she is now back in the fourth round , where she will face 11th - seeded je \#\#lena jan \#\#kovic of serbia , a 6 - 3 , 6 - 4 winner over [unused0] victoria az \#\#are \#\#nka [unused1] \textbf{of} [unused2] belarus [unused3] . [SEP]      &         /people/person/nationality     \\ \hline
      [CLS] [unused2] boston [unused3] what : a two - bedroom condo how much : \$ 59 \#\#9 , 000 per square foot : \$ 83 \#\#6 located in the [unused0] back bay [unused1] area of the city , this 71 \#\#6 - square - foot condo has views from the apartment and its private roof deck of the charles river , one block away . [SEP] seven years ago , when nad \#\#er tehran \#\#i and monica ponce de leon , partners at office da , an architecture firm in [unused2] boston [unused3] , were asked to reno \#\#vate a five - story town house in the [unused0] back bay [unused1] \textbf{neighborhood} , they faced a singular design challenge . [SEP] far more inviting is first church in [unused2] boston [unused3] , in [unused0] back bay [unused1] , which replaced a gothic building that burned in 1968 . [SEP]      & /location/neighborhood/neighborhood\_of   \\ \hline
      [CLS] [unused0] michael sm \#\#uin [unused1] , a choreographer who worked for major ballet companies and led his own , marshal \#\#ing eclectic dance forms , robust athletic \#\#ism and striking theatrical \#\#ity to create works that appealed to broad audiences , \textbf{died} yesterday in [unused2] san francisco [unused3] . [SEP] & /people/deceasedperson/place\_of\_death \\ \hline
    [CLS] [unused2] steve new \#\#comb [unused3] , a [unused0] powers \#\#et [unused1] \textbf{founder} and veteran of several successful start - ups , said his company could become the next google . [SEP] & /business/company/founders \\ \hline
       
    \end{tabular}}
    \caption{Attention analysis on a few random correctly predicted non-NA triples on NYT-10m test set. The highest attention-scored token (excluding entity mentions and special markers and stop words) are present in bold. [unused0], [unused1] denote the start and end head entity markers. [unused2], [unused3] denote the start and end tail entity markers.}
    \label{tab:att_viz}
\end{table}

\section{Performance vs Length of test passages}
\label{sec:appendix_len}

Our instance aggregation scheme truncates the passage if the number of tokens exceed the maximum number of tokens allowed by the encoder. In such cases, one would assume that the our model is not suited for cases where the number of instances present in a bag is very large. To test this hypothesis, we divide the non-NA bags, $(e_1,e_2)$, present in the NYT-10m data into 6 bins based on the number of tokens present in $P(e_1,e_2)$ (after tokenized using BERT). We then compare the performance with CIL on examples present in each bin. The results in figure \ref{fig:bin_main} indicate that a) our model beats CIL in each bin-size b) the performance trend across different bins is the same for both models. This trend is continued even for passages where the number of tokens present exceed the maximum number of tokens allowed for BERT (i.e. 512). This results indicate that 512 tokens provide sufficient information for correct classification of a triple. Moreover, models using intra-bag attention aggregation scheme fix the number of instances sampled from the bag in practice. For CIL, the best performing configuration uses a bag-size of 3. This analysis therefore indicates that our model doesn't particularly suffer a drop in performance on large bags when compared with other state-of-the-art models. 

\section{Entity Permutation Test}
\label{sec:appendix_ood}



To understand how robust our trained model would be to changes in the KB, we design the entity permutation test (inspired by \citet{ribeiro2020beyond}). An ideal DS-RE model should be able to correctly predict the relationship between an entity pair by understanding the semantics of the text mentioning them. Since DS-RE models under the multi-instance multi-label \cite{surdeanu-etal-2012-multi} (MI-ML) setting are evaluated on bag-level, it might be the case that such models are simply memorizing the KB on which they are being trained on. 

To test this hypothesis, we construct a new test set (in fact, 5 such sets and report average over those 5) using NYT-10m by augmenting its KB. Let $B(e_1,e_2)$ denote a non-NA bag already existing in the test set of the dataset. We augment this bag to correspond to a new entity-pair (which is not present in the combined KB of all three splits of this dataset). The augmentation can be of two different types: replacing $e_1$ with $e_1'$ or replacing $e_2$ with $e_2'$. We restrict such augmentations to the same type (i.e the type of $e_i$ and $e_i'$ is same for $i={1,2}$). For each non-NA entity pair in the test set of the dataset, we select one such augmentation and appropriately modify each instance in $B(e_1,e_2)$ to have the new entity mentions. We note that since each instance in NYT-10m is manually annotated and since our augmentation ensures that the type signature is preserved, the transformation is label preserving. For the NA bags, we use the ones already present in the original split. This entire transformation leaves us with an augmented test set, having same number of NA and non-NA bags as the original split. The non-NA entity pairs are not present in the KB on which the model is trained on. 

\section{More Analysis on DiS-ReX}
\label{sec:appendix_disrex}
\subsection{Relation-wise F1 scores}
To show how our model performs on each relation label compared to other competitive baselines, we present relation-wise F1 scores on DiS-ReX in table  \ref{tab:disrex_classwise}.
\begin{table}[h!]
\centering
\begin{tabular}{@{}lrrl@{}}
\toprule
\textbf{Relation} & \textbf{\mpare} & \textbf{mBERT-MNRE} & \textbf{mBERT-Avg} \\ \midrule
\href{http://dbpedia.org/ontology/birthPlace}{http://dbpedia.org/ontology/birthPlace} &	\textbf{77.5} &	\underline{75.3} &	74.9 \\
\href{http://dbpedia.org/ontology/associatedBand}{http://dbpedia.org/ontology/associatedBand}	& \textbf{77.9} &	70.9 & \underline{74.7}        \\ 
\href{http://dbpedia.org/ontology/director}{http://dbpedia.org/ontology/director}	& \textbf{88.4}	& 83.2	& \underline{85.5} \\
\href{http://dbpedia.org/ontology/country}{http://dbpedia.org/ontology/country}	& \textbf{88.4} &	\underline{86}	& 85.2 \\
\href{http://dbpedia.org/ontology/deathPlace}{http://dbpedia.org/ontology/deathPlace}	& \textbf{71.0}	& \underline{67.3} & 65.5 \\
\href{http://dbpedia.org/ontology/nationality}{http://dbpedia.org/ontology/nationality}	& \textbf{70.4}	& 67.7	& \underline{68.7} \\
\href{http://dbpedia.org/ontology/location}{http://dbpedia.org/ontology/location}	& \textbf{74.2}	& \underline{70.5}	& 67.5 \\
\href{http://dbpedia.org/ontology/related}{http://dbpedia.org/ontology/related}	& \textbf{78.9}	& \underline{75.5} &	73.2 \\
\href{http://dbpedia.org/ontology/isPartOf}{http://dbpedia.org/ontology/isPartOf} &	\textbf{74.8} & \underline{68.6} &	64.7 \\
\href{http://dbpedia.org/ontology/influencedBy}{http://dbpedia.org/ontology/influencedBy}	& \underline{57.7}	& \textbf{58.4}	& 57.4 \\
\href{http://dbpedia.org/ontology/starring}{http://dbpedia.org/ontology/starring}	& \textbf{87.5}	& \underline{86.1}	& 83.9 \\
\href{http://dbpedia.org/ontology/headquarter}{http://dbpedia.org/ontology/headquarter} &	\textbf{74.0}	& \underline{70.7} & 66.7 \\
\href{http://dbpedia.org/ontology/successor}{http://dbpedia.org/ontology/successor}	& \textbf{74.2}	& \underline{71.8}	& 71.3 \\
\href{http://dbpedia.org/ontology/bandMember}{http://dbpedia.org/ontology/bandMember}	& \textbf{76.2}	& \underline{74.6}	& 74.3 \\
\href{http://dbpedia.org/ontology/producer}{http://dbpedia.org/ontology/producer} &	\textbf{56.7}	& \underline{53.6} &	48.5 \\
\href{http://dbpedia.org/ontology/recordLabel}{http://dbpedia.org/ontology/recordLabel}	& \textbf{90.5}	& \underline{86.9} & 86.1\\
\href{http://dbpedia.org/ontology/city}{http://dbpedia.org/ontology/city} &	\textbf{83.2}	& \underline{78.8} &	77.6 \\
\href{http://dbpedia.org/ontology/influenced}{http://dbpedia.org/ontology/influenced} &	\underline{56.3} &	\textbf{61.9}	& 51.5 \\
\href{http://dbpedia.org/ontology/author}{http://dbpedia.org/ontology/author} &	\textbf{81.6}	& 78.2	& \underline{80.5} \\
\href{http://dbpedia.org/ontology/team}{http://dbpedia.org/ontology/team}	& \textbf{84.8} &	\underline{82.5} &	78.6 \\
\href{http://dbpedia.org/ontology/formerBandMember}{http://dbpedia.org/ontology/formerBandMember} &	56.4	& \textbf{57.4} &	\underline{56.5} \\
\href{http://dbpedia.org/ontology/state}{http://dbpedia.org/ontology/state}	& \textbf{86.9}	& \underline{83.9} & 82.4 \\
\href{http://dbpedia.org/ontology/region}{http://dbpedia.org/ontology/region}	& \textbf{84.8}	& \underline{80.4} &	78.8 \\
\href{http://dbpedia.org/ontology/subsequentWork}{http://dbpedia.org/ontology/subsequentWork} &	\textbf{74.1}	& \underline{72.4}	& 69.6 \\
\href{http://dbpedia.org/ontology/department}{http://dbpedia.org/ontology/department} &	\textbf{96.4}	& 95.4	& \underline{95.5} \\
\href{http://dbpedia.org/ontology/locatedInArea}{http://dbpedia.org/ontology/locatedInArea}	& \textbf{76.4}	& \underline{72.5}	& 72.3 \\
\href{http://dbpedia.org/ontology/artist}{http://dbpedia.org/ontology/artist}	& \textbf{80.8}	& 77.2	& \underline{78.6} \\
\href{http://dbpedia.org/ontology/hometown}{http://dbpedia.org/ontology/hometown}	& \textbf{78.8}	& 73.6	& \underline{73.7} \\
\href{http://dbpedia.org/ontology/province}{http://dbpedia.org/ontology/province}	& \textbf{82.1}	& \underline{79.2}	& 78.2 \\
\href{http://dbpedia.org/ontology/riverMouth}{http://dbpedia.org/ontology/riverMouth}	& \textbf{77.2}	& \underline{72.4}	& 71.9 \\
\href{http://dbpedia.org/ontology/locationCountry}{http://dbpedia.org/ontology/locationCountry}	& \textbf{66.9}	& 62.5	& \underline{64.2} \\
\href{http://dbpedia.org/ontology/predecessor}{http://dbpedia.org/ontology/predecessor} &	\underline{67.3}	& \textbf{68.1}	& 62 \\
\href{http://dbpedia.org/ontology/previousWork}{http://dbpedia.org/ontology/previousWork} &	\underline{68.6} &	\textbf{69.6}	& 65.5 \\
\href{http://dbpedia.org/ontology/capital}{http://dbpedia.org/ontology/capital}	& \textbf{68.6}	& 55.1	& \underline{58} \\
\href{http://dbpedia.org/ontology/leaderName}{http://dbpedia.org/ontology/leaderName}	& \textbf{78.4}	& \underline{70.4}	& 63.3 \\
\href{http://dbpedia.org/ontology/largestCity}{http://dbpedia.org/ontology/largestCity}	& \textbf{65.7}	& \underline{59.1}	& 48.6 \\
\bottomrule
\end{tabular}
\caption{Relation-wise F1 scores on DiS-Rex. Bold and underline represent best and second best models respectively on a class. Our model consistently beats the other 2 models in 31 out of 36 relation classes, thus showing how strong our approach is for the multilingual setting. }
\label{tab:disrex_classwise}
\end{table}
\subsection{Language-wise AUC scores}
We compare the performance of our model compared to other baselines on every language in DiS-ReX. For this, we partition the test data into language-wise test sets i.e. containing instances of only a particular language. The results are presented in table \ref{tab:langwise}. We observe that the order of performance across languages is consistent for all models including ours i.e. German < English < Spanish < French. Further we observe that our model beats the second best model by an AUC ranging from 3 upto 4 points on all languages.

\begin{table}[h!]
\centering
\begin{tabular}{@{}lrrrl@{}}
\toprule
  \textbf{Model} & \textbf{English} & \textbf{French} & \textbf{German} & \textbf{Spanish} \\ 
 \hline
 \mpare & \textbf{83.2} & \textbf{86.8} & \textbf{81.7} & \textbf{85.3} \\ 
 mBERT-Avg & \underline{79.9} & \underline{83.1} & \underline{77.7} & \underline{82.1} \\
 mBERT-MNRE & 79.6 & 82.2 & 75.5 & 81.6 \\
 
 \bottomrule
\end{tabular}
\caption{Language-wise AUC comparison of our model v/s baseline models.}
\label{tab:langwise}
\end{table}

\subsection{Do multilingual bags improve performance?}
\begin{figure}[h!]
\caption{AUC vs number of languages in a bag in DiS-ReX test set}
\label{fig:disrex_lang}
\centering
\includegraphics[width=8cm, height=6cm]{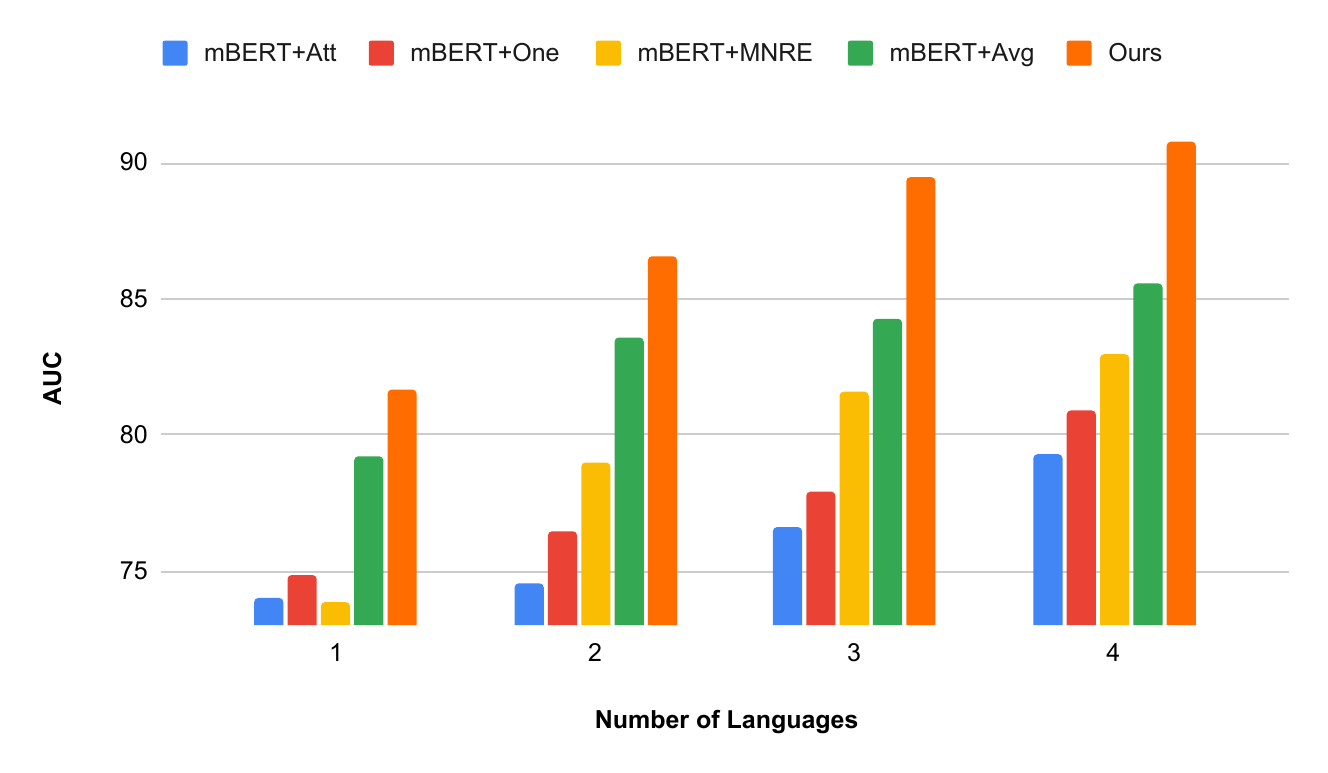}
\end{figure}

To understand whether the currently available aggregation schemes (including ours) are able to benefit from multilingual bags or not, we conduct an experiment where we only perform inference on test-set bags that contain instances from all four languages. In the multilingual case, the \emph{passage} constructed during the \emph{Passage Summarization} step will contain multiple sentences of different languages. To understand whether such an input allows improves (or hampers) the performance, we devise an experiment where we perform inference by removing sentences from any one, two or three languages from the set of bags containing instances of all four languages. There are roughly 1500 bags of such kind. Note that removing any $k$ languages ($k<=3$) would result in $4 \choose k$ different sets and we take average of AUC while reporting the numbers. The results are presented in figure \ref{fig:disrex_lang}. 

We observe that in all aggregation schemes, AUC increases with increase in number of languages of a multilingual bag. \mpare\ consistently beats the other models in each scenario, indicating that the encoding of a multilingual passage and attention-based summarization over multilingual tokens doesn't hamper the performance of a DS-RE model with increasing no. of languages.

\section{Negligible effect of random ordering}
\label{sec:appendix_random}
Since we order the sentences randomly into a passage to be encoded by BERT, this may potentially cause some randomness in the results. However, we hypothesize that the BERT encoder must also be getting fine-tuned to treat the bag as a set (and not a sequence) of sentences when being trained with random ordering technique. And as a result, it's performance must be agnostic to the order of sentences it sees in a passage during inference. To validate this, we perform 20 inference runs of our trained model with different seeds i.e. the ordering of sentences is entirely random in each run. We measure mean and standard deviation for each dataset as listed in table \ref{tab:diff_seed}. We observe negligible standard deviation in all metrics. A minute variation in Macro-F1 or P@M metrics may be attributed to the fact that these are macro-aggregated metrics and a variation in performance over some data points may also affect these to some extent.  

\begin{table}[htbp]
  \centering
  \small{\begin{tabular}{*{9}{l}}
    \toprule
    & \multicolumn{2}{c}{\textbf{NYT-10m}} & \multicolumn{2}{c}{\textbf{NYT-10d}} 
      & \multicolumn{2}{c}{\textbf{Wiki-20m}} & \multicolumn{2}{c}{\textbf{DiS-ReX}} \\
    \cmidrule(lr){2-3}
    \cmidrule(lr){4-5}
    \cmidrule(lr){6-7}
    \cmidrule(lr){8-9}
     & \textbf{AUC} & \textbf{M-F1} & \textbf{AUC} & \textbf{P$@$M} & \textbf{AUC}  & \textbf{M-F1} & \textbf{AUC}  & \textbf{M-F1} \\
    \midrule
     & 62.11 & 38.35 & 53.49 & 84.82 & 91.41	& 83.87 & 87.03 & 	76.01 \\
     & 62.11 & 38.44 & 53.43 & 84.72  & 91.41 & 83.88 & 87.06 & 	76.18 \\
     & 62.18 & 38.27 & 53.49	& 84.69  & 91.41 & 83.85 & 87.0 & 	76.04 \\
     & 62.11 & 38.32 & 53.45	& 84.56  & 91.42 & 83.88 & 86.98 & 	75.93 \\
     & 62.12 & 38.34 & 53.64	& 84.62  & 91.43 & 84.04 & 87.03 & 	76.03 \\
     & 62.25 & 38.46 & 53.6	& 84.73  & 91.42 & 83.82 & 87.04 & 	76.07 \\
     & 62.16 & 38.54 & 53.54	& 85.18  & 91.42 & 83.81 & 87.01 & 	76.0 \\
     & 62.2 & 38.68 & 53.45 &	84.57  & 91.41	& 83.91 & 86.99 & 	75.98 \\
     & 62.22 & 38.27 & 53.43	& 84.4  & 91.42 & 83.83 & 87.06 & 	76.2 \\
     & 62.19	 & 38.47 & 53.47	& 84.68   & 91.41 & 83.81 & 87.02 & 	76.06\\
     & 62.22 & 38.43 & 53.45	& 84.51   & 91.41 &	83.85 & 87.03 & 75.99\\
     & 62.13 & 38.4 & 53.5	& 85.18   & 91.41	& 83.85 & 87.06 & 	76.14\\
     & 62.21 & 38.3 & 53.58	& 85.23   & 91.42 &	83.87 & 87.02 & 	75.96\\
     & 62.18 & 38.15 & 53.4	& 84.51  & 91.43	& 83.91 & 87.01 & 	75.97\\
     & 62.21 & 38.51 & 53.44 &	84.54   & 91.41	& 83.88 & 87.04 & 	76.1\\
     & 62.2 & 38.34 & 53.53	& 84.51   & 91.41	& 83.91 & 87.03 &	76.04\\
     & 62.13 & 38.29 & 53.61	& 84.56   & 91.43	& 83.96 & 87.02 & 	76.05\\
     & 62.23 & 38.63 & 53.46 &	84.79   & 91.41	& 83.81 & 87.04 & 	76.13\\
     & 62.19 & 38.3 & 53.42	& 84.46   & 91.41	& 83.85 & 87.03 &	75.96\\
     & 62.29 & 38.36 & 53.47 & 85.07  &  91.42	& 83.87 & 87.01 &	76.01\\
     
    \bottomrule
    \textbf{Average} & 62.18 & 38.39 & 53.49 & 84.71 & 91.42 & 83.87 & 87.03 & 76.01 \\
    \textbf{Std-Dev} & 0.05 & 0.13 & 0.07 & 0.25 & 0.01 & 0.06 & 0.01 & 0.07\\
    \textbf{Std-Dev(\%)}  & 0.08 & 0.34 & 0.13 & 0.3 & 0.01 & 0.07 & 0.01 & 0.1 \\
  \end{tabular}}
  \caption{We perform 20 inference runs with random seeds of our trained model on each dataset and report the mean and standard deviation. All numbers have been rounded upto second decimal place. We observe negligible stdandard deviation in all metrics on all datasets thus validating our hypothesis that the model learns to treat a bag of sentences as a set (and not a sequence) of sentences treating any random order almost alike. Note that the results presented in main paper are for inference done with same seed value with which the model has been trained. However, in current analysis we select random seed values at inference (irrespective of the one with which it was trained).}

\label{tab:diff_seed}
\end{table}
\end{document}